\documentclass{article}

\usepackage[preprint]{corl_2026} 
\usepackage[utf8]{inputenc} 
\usepackage[T1]{fontenc}    
\usepackage{hyperref}       
\usepackage{url}            
\usepackage{booktabs}       
\usepackage{amsfonts}       
\usepackage{nicefrac}       
\usepackage{microtype}      
\ifdefined\XeTeXversion
  \catcode`’=\active
  \def’{'}
\else
  \DeclareUnicodeCharacter{2019}{'}
\fi
\usepackage[table]{xcolor}  
\usepackage{amsmath}
\usepackage{multirow}
\usepackage{booktabs}  
\usepackage{pifont}    
\usepackage{bbding}    
\usepackage{graphicx}
\usepackage{caption}
\usepackage{wrapfig}
\usepackage{enumitem}
\usepackage{xspace}
\usepackage{caption}
\usepackage{enumitem}
\usepackage{booktabs}
\usepackage{float}
\definecolor{mygreen}{HTML}{A4C244} 
\definecolor{myred}{HTML}{C00000}   
\definecolor{myblue}{HTML}{6495ED}  
\definecolor{mygray}{HTML}{666666}  
\definecolor{rowgray}{HTML}{EBEBEB} 

\title{AdvDex: Learning Dexterous Manipulation from Human Demonstrations via Joint-Aligned Actions and Adversarial Learning}

\author{
 Zhiyue Zhao$^{1,2}$\quad
 Jingyi Wu$^{4}$\quad
 Hairuo Liu$^{3,2}$\quad
 Mingyu Liu$^{1,2}$\quad \\
 \textbf{Liyang Li}$^{1}$\quad
 \textbf{Hengdi Zhang}$^{5}$\quad
 \textbf{Tong He}$^{2}$\quad
 \textbf{Zhengxue Cheng}$^{3}$\thanks{Corresponding author.}\quad\\
 $^{1}$Zhejiang University \quad
 $^{2}$Shanghai Innovation Institute \quad \\
 $^{3}$Shanghai Jiao Tong University \quad
 $^{4}$Fudan University \quad
 $^{5}$Paxini Tech
}

\begin{document}
\maketitle

\begin{abstract}
Dexterous manipulation is a fundamental capability for embodied intelligence, but scaling it remains difficult because robot demonstrations are expensive to collect and action spaces vary across embodiments. Policies trained on heterogeneous data can also entangle task-relevant visual cues with embodiment-specific appearance, limiting cross-embodiment generalization. We present AdvDex, a unified Vision-Language-Action framework for learning dexterous manipulation from human and robot demonstrations. First, we introduce OmniShare, a large-scale multimodal dataset of human manipulation demonstrations that provides high-quality kinematic supervision and tactile measurements while reducing reliance on robot teleoperation. Second, we propose the Joint-Aligned Action Space (JAAS), a canonical action representation comprising an $\mathrm{SE}(3)$ wrist pose and 15 finger joints, thereby functionally aligning human hands, dexterous robot hands, and parallel grippers. Finally, we use domain-adversarial learning to reduce embodiment-specific information in the learned visual representation. Experiments on hand-action prediction and real-world dexterous manipulation show consistent improvements over baselines, effective zero-shot human-to-robot skill transfer, generalization to unseen objects and environments, and data-efficient few-shot adaptation.
\end{abstract}

\section{Introduction}
\begin{figure*}[t]
    \centering
    \includegraphics[width=\textwidth]{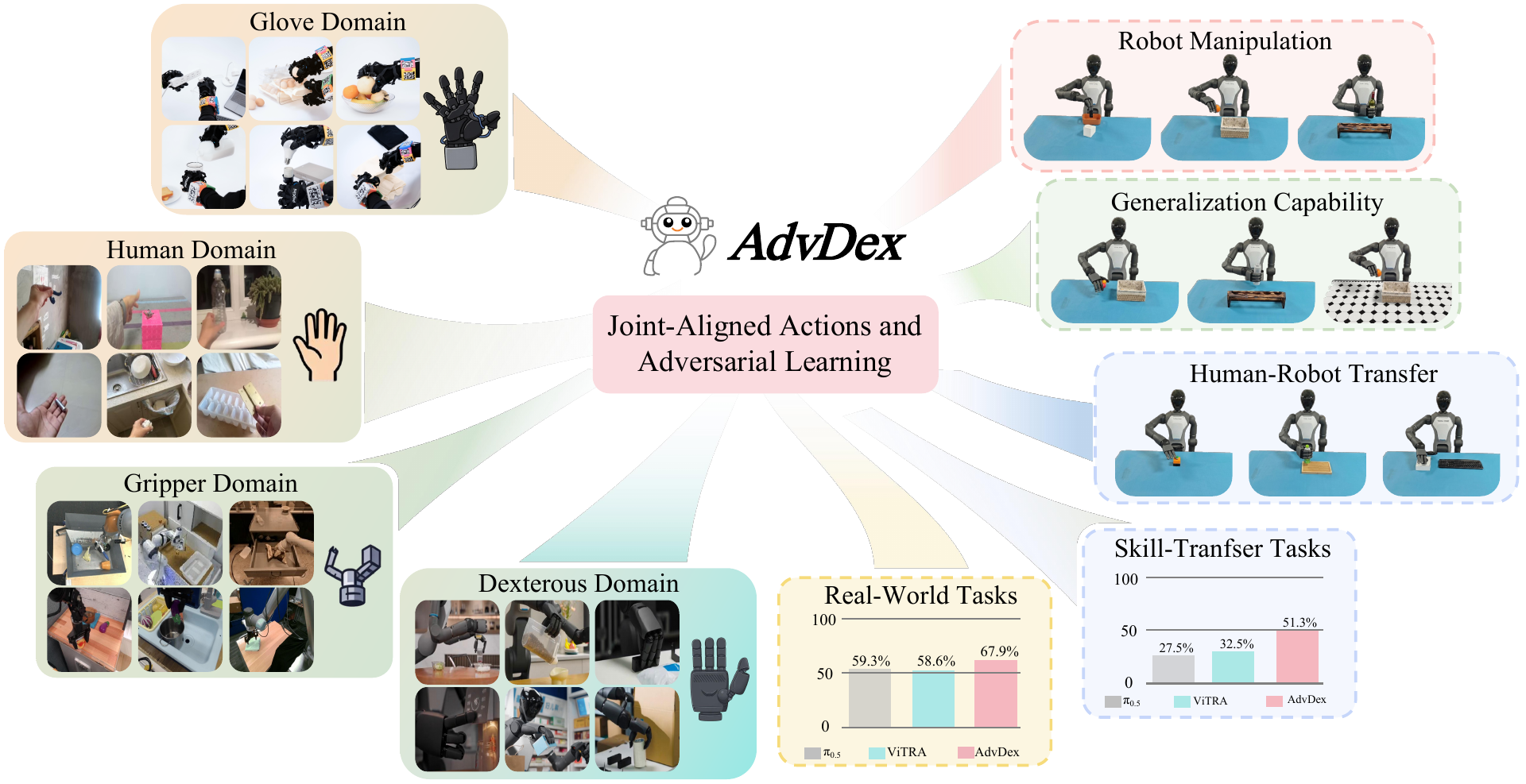}
    \caption{
        \textbf{AdvDex.}
        A unified Vision-Language-Action framework for scalable cross-embodiment
        dexterous manipulation. By aligning hardware kinematics through a shared
        action space and reducing embodiment-specific visual information through
        domain-adversarial learning, AdvDex supports zero-shot generalization to
        novel objects and unseen environments.
    }
    \label{fig:teaser}
\end{figure*}

In recent years, learning robotic manipulation policies from teleoperated demonstrations has advanced rapidly \cite{rajeswaran2018learning, chi2023diffusion, cheng2024opentelevision,liu2025gae, huang2025notvla}. However, collecting large-scale robot datasets remains expensive and labor-intensive \cite{khazatsky2024droid, o2024open}, creating a persistent data bottleneck for general-purpose manipulation. While most robotic foundation policies focus on parallel-jaw grippers \cite{o2024open, team2024octo}, complex tool use can be difficult or infeasible for such simple end-effectors. Because dexterous robot hands share similar actuation patterns with human hands, human data provide a promising alternative: it is abundant, easier to acquire, and rich in manipulation behaviors \cite{liu2022hoi4d, hoque2025egodex}. Advances in wearable sensing have further motivated the use of human motion data for robot-policy co-training and pre-training \cite{engel2023projectaria, kareer2024egomimic,yuan2025motiontrans,liu2026stamo}.

Despite this progress, dexterous manipulation data remain fragmented across hardware platforms. Robot hands such as Wuji, Xhand, Shadow, and Paxini DexH13 differ in kinematic structure, degrees of freedom, joint constraints, and appearance, making demonstrations collected on one platform difficult to reuse on another. These morphological differences leave human and robot dexterous manipulation without a shared action representation \cite{yuan2025hermes, li2025maniptrans}. A second challenge arises in visual representation learning: shared visual encoders can entangle task-relevant spatial information with embodiment-specific appearance cues, limiting cross-embodiment generalization and zero-shot transfer~\cite{liu2025egozero,yuan2025motiontrans}.

These challenges are closely coupled. Simply pooling human and robot trajectories does not make their actions directly comparable, while action-space alignment alone cannot prevent the visual backbone from exploiting embodiment identity as a shortcut. A transferable policy must therefore address both kinematic alignment and embodiment-invariant representation learning.

To address these challenges, we present \textbf{AdvDex}, a unified Vision-Language-Action (VLA) framework for cross-embodiment dexterous manipulation. We first introduce \textbf{OmniShare}, a multimodal dataset of human manipulation demonstrations comprising over 100k trajectories, 500 tasks, and 700 objects. Captured with a microsecond-synchronized sensor suite, OmniShare provides high-quality kinematic supervision and tactile measurements while reducing reliance on robot teleoperation. Next, we propose the \textbf{Joint-Aligned Action Space (JAAS)}, a canonical representation with a shared $\mathrm{SE}(3)$ wrist pose and 15 finger joints. JAAS defines functional correspondences across 51-DoF MANO human states~\cite{Romero2017EmbodiedH}, 19-DoF dexterous hands, and systems with a 7-DoF arm and parallel gripper. Finally, we use \textbf{domain-adversarial learning} with a Gradient Reversal Layer (GRL) to suppress embodiment-specific appearance cues and learn more transferable visual features.

In summary, our primary contributions are threefold:
\begin{itemize}[leftmargin=*]
    \item \textbf{OmniShare Dataset:} We introduce a large-scale multimodal dataset covering over 500 tasks and 700 objects, providing high-quality human supervision while reducing reliance on robot teleoperation.
    \item \textbf{Joint-Aligned Action Space:} We propose a canonical representation comprising an $\mathrm{SE}(3)$ wrist pose and 15 finger joints that provides a shared action interface for human hands, dexterous robot hands, and parallel grippers.
    \item \textbf{Domain-Adversarial Learning:} We integrate a VLM-DiT architecture with a Gradient Reversal Layer to reduce embodiment-specific information in the learned visual representation.
\end{itemize}

Together, these components form a unified framework for cross-embodiment manipulation. Physical experiments show that AdvDex consistently outperforms the evaluated baselines and supports effective zero-shot policy transfer and few-shot adaptation in unseen environments.

\section{Related Work}
\subsection{Vision-Language-Action Models}
Robotic Vision-Language-Action (VLA) models \cite{brohan2022rt,team2024octo,zitkovich2023rt,o2024open,black2410pi0,liu2024rdt,wen2025dexvla, wang2025vq, su2026world, liu2026perfect} learn diverse tasks through large-scale pre-training. Although visual representations can benefit from abundant image and video data, scaling action supervision remains difficult because real-robot data are expensive to collect \cite{o2024open, khazatsky2024droid}. Recent approaches address this limitation with auxiliary data \cite{ji2025robobrain,liu2024segment}, simulation \cite{deng2025graspvla}, and latent representations extracted from other data sources \cite{yelatent2025,chen2024igor,bjorck2025gr00t,wu2023unleashing}. Building on architectures such as Diffusion Policy \cite{chi2023diffusion} and $\pi_0$ \cite{black2410pi0}, our work addresses this bottleneck by scaling action-space pre-training with high-quality human demonstrations.

\subsection{Learning from Human Demonstrations}
Human demonstrations offer an abundant alternative to teleoperation \cite{hoque2025egodex, liu2022hoi4d, damen2022epic}. While early methods extract visual priors or trajectories from videos \cite{ma2023vip,nair2023r3m,yang2024spatiotemporal,bahl2023affordances,mandikal2022dexvip,bharadhwaj2024track2act,wen2023any,mendonca2023structured,chen2024igor}, recent work uses VR devices \cite{engel2023projectaria, chen2024arcap, hoque2025egodex} for end-to-end motion supervision \cite{lepert2025phantom}. Bridging the morphological gap often requires sim-to-real retargeting \cite{yuan2025hermes,li2025maniptrans}, human-in-the-loop corrections \cite{wang2024dexcap}, or alignment fine-tuning \cite{yang2025egovla, luo2025being, bi2025h, liu2025egozero}. Our framework instead maps human data directly into a canonical action space, enabling joint training across human and robot embodiments.

\subsection{Dexterous Hand Manipulation}
Dexterous manipulation has progressed from analytic control \cite{mordatch2012contact,bai2014dexterous,kerr1986analysis} and simulation-based reinforcement learning \cite{akkaya2019solving,andrychowicz2020learning, qi2023general,lin2025learning} to real-world imitation learning through teleoperation \cite{rajeswaran2018learning, cheng2024opentelevision} and human motion data \cite{handa2020dexpilot,wang2023mimicplay,wang2024dexcap,qiu2025humanoid}. Even with language conditioning \cite{zhong2025dexgraspvla,hu2025video}, many policies remain hardware-specific or focus on narrow tasks such as static grasping \cite{fang2025anydexgrasp,zhong2025dexgraspvla,yuan2024learning,si2024difftactile,ding2024preafford}. In contrast, we map scalable human demonstrations into a unified Joint-Aligned Action Space, enabling pre-training of a shared policy across human and robot embodiments.

\section{Method}
\subsection{OmniShare Dataset}

\begin{figure}[htbp]
    \centering
    \begin{minipage}[t]{0.6\linewidth}
        \vspace{0pt} %
        \centering
        \includegraphics[width=\linewidth]{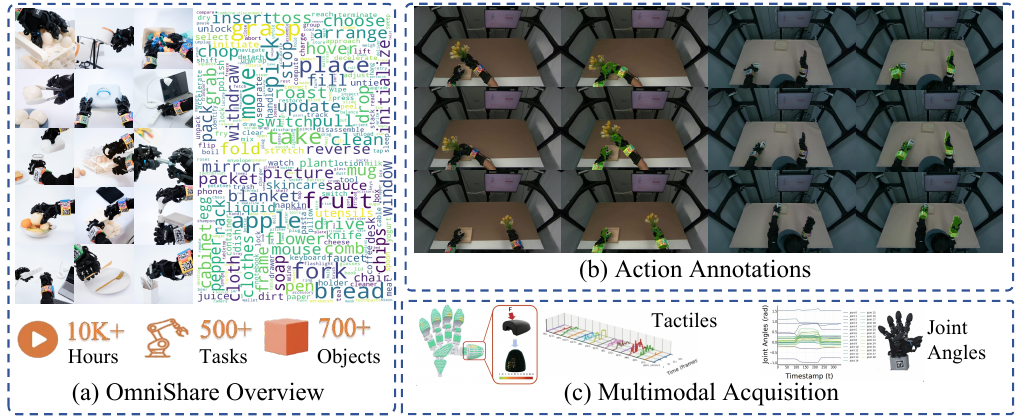}
    \caption{\textbf{Overview of OmniShare.} The dataset contains over 10k hours of videos covering 500 tasks and 700 objects.}
        \label{fig:data}
    \end{minipage}
    \hfill %
    \begin{minipage}[t]{0.36\linewidth}
        \vspace{0pt} %
        \centering
        \captionof{table}{Comparison of 3D bimanual motion datasets.}
        \label{tab:dataset_comparison}
        \vspace{2mm} 
        \resizebox{\linewidth}{!}{
        \begin{tabular}{l | c c c c}
            \toprule
            Name & trajs & views & obj. & Text \\
            \midrule
            AssemblyHands \cite{ohkawa2023assemblyhands}  & 62 & 12 & / & Sparse \\
            Ego-Exo4D \cite{grauman2024ego}      & / & 5-6 & / & Sparse \\
            HOI4D \cite{liu2022hoi4d}         & 4k & 1 & \textbf{800} & Sparse \\
            ARCTIC \cite{Fan2022ARCTICAD}         & 339 & 9 & 11 & None \\
            TACO \cite{liu2024taco}         & 2.3k & 13 & 196 & Sparse \\
            OakInk2 \cite{zhan2024oakink2}      & 2.8k & 4 & 75 & Sparse \\
            HOT3D \cite{banerjee2025hot3d}       & 4.1k & 2-3 & 33 & None \\
            GigaHands \cite{fu2025gigahands}         & 13.9k & 51 & 417 & Dense \\
            OmniShare (Ours)   & \textbf{168k} & \textbf{14} & 721 & Dense \\
            \bottomrule
        \end{tabular}
        }
    \end{minipage}
\end{figure}

\subsubsection{Dataset Overview}
We introduce \textbf{OmniShare}, a diverse dexterous manipulation dataset comprising over 100k trajectories across five real-world domains (Fig.~\ref{fig:data}a). It covers more than 500 manipulation tasks and 700 objects, with substantial variation in object geometry and materials to support policy generalization.

\subsubsection{Data Collection System}
We collect demonstrations using a microsecond-synchronized multimodal sensor suite~\cite{Fan2026RoboPaintFH}. The operator wears a data glove equipped with 29 magnetic rotary encoders for sub-degree hand kinematics and a Hall-effect tactile array for measuring normal contact forces. The synchronized system records hand motion, contact, and multi-view scene observations on a shared timeline, providing direct articulation and contact signals together with contextual information about instructions, objects, and surrounding geometry.

\subsubsection{Data Processing Pipeline}
We first estimate 6D wrist and object poses using ArUco markers and FoundationPose. A physics-aware optimization then retargets the raw hand states into a canonical MANO representation by jointly minimizing kinematic and tactile discrepancies. A distance-aware decay function modulates the tactile signals (Fig.~\ref{fig:data}c) to preserve contact timing and grasp-force variation.

Each demonstration retains its original visual observation and language instruction, while its motion is represented in the same intermediate hand space before being mapped into JAAS. This enables human trajectories to provide action supervision across heterogeneous embodiments without requiring robot-specific labels for every human joint.

\begin{wrapfigure}{r}{0.43\textwidth}
    \vspace{-15pt}
    \centering
    \includegraphics[width=\linewidth]{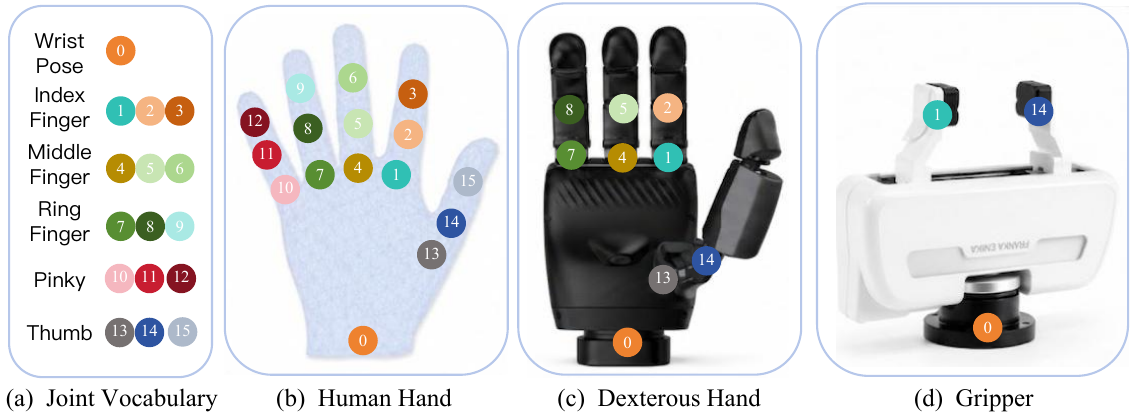}
    \caption{\textbf{Joint-Aligned Action Space.} A canonical $\mathrm{SE}(3)$ wrist and 15-joint representation shared across embodiments.}
    \label{fig:mapping}
    \vspace{-10pt}
\end{wrapfigure}

\subsection{Unified Cross-Embodiment Framework}

\subsubsection{Joint-Aligned Action Space}
Hardware action-space heterogeneity remains a major bottleneck in cross-embodiment manipulation. To address this, we propose the Joint-Aligned Action Space (JAAS), a canonical action representation shared across human hands, dexterous robot hands, and parallel grippers. JAAS maps different embodiments into a common joint vocabulary consisting of an $\mathrm{SE}(3)$ wrist pose (3D translation and continuous 3D rotation) and 15 finger joints, with three 3-DoF Euler joints per finger.

Specifically, human demonstrations represented by the MANO model~\cite{Romero2017EmbodiedH} are mapped to designated canonical slots (Fig.~\ref{fig:mapping}b). The joints of 19-DoF dexterous hands are similarly assigned to functionally corresponding slots (Fig.~\ref{fig:mapping}c). For systems with a 7-DoF arm and parallel gripper, the 1-DoF jaw action is mapped to two canonical finger slots according to functional correspondence (Fig.~\ref{fig:mapping}d). The shared $\mathrm{SE}(3)$ wrist pose captures the motion of the end-effector, providing a unified action interface across heterogeneous embodiments.

JAAS relies on functional correspondence rather than identical anatomy. Unavailable canonical slots are masked during action-loss computation, while active slots retain consistent semantics across embodiments. This enables a single action expert to learn from heterogeneous kinematic structures while preserving embodiment-specific feasibility at execution time.

\begin{figure*}[t]
    \centering
    \includegraphics[width=\textwidth]{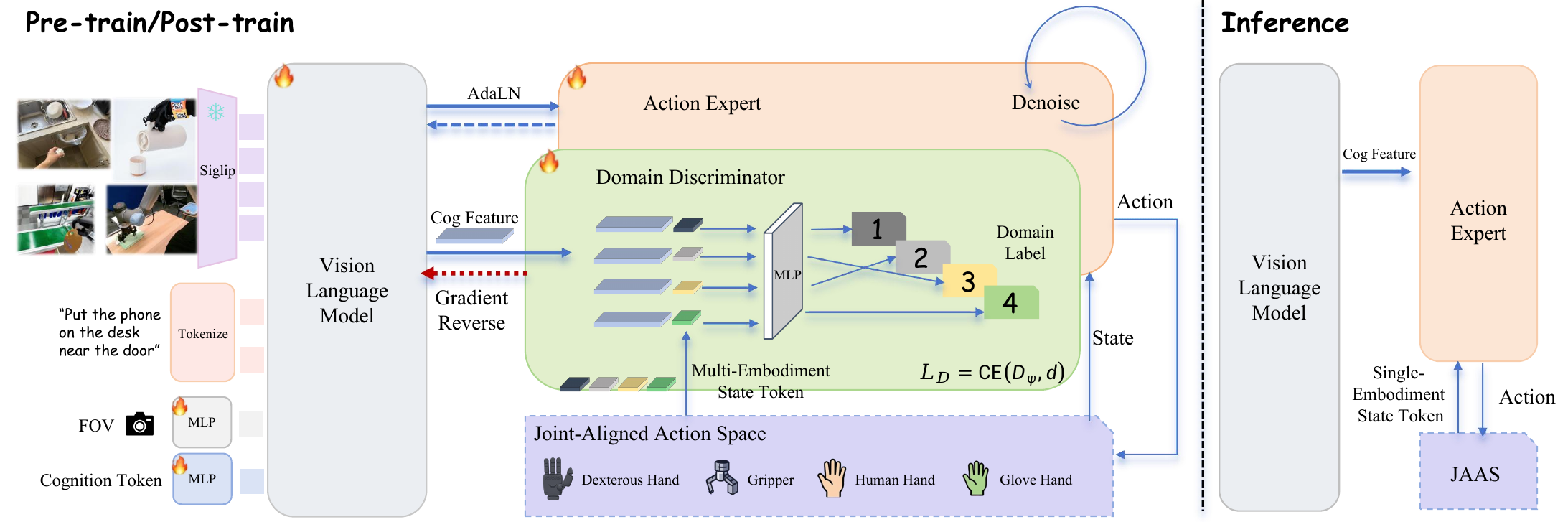}
    \caption{\textbf{Overview of the Domain-Adversarial VLA.} A domain discriminator connected to the Vision-Language Model through a Gradient Reversal Layer encourages the extracted cognition tokens to become more domain-invariant before they condition the Diffusion Transformer for joint-aligned action prediction.}
    \label{fig:pipeline}
\end{figure*}

\subsubsection{Domain-Adversarial Learning}

While JAAS addresses kinematic discrepancies, shared visual encoders can still entangle task-relevant spatial information with embodiment-specific appearance cues. We therefore introduce a domain-adversarial architecture (Fig.~\ref{fig:pipeline}) comprising a Vision-Language Model (VLM) backbone $E_{\theta}$, a Diffusion Transformer (DiT) action expert $P_{\phi}$, and a domain discriminator $D_{\psi}$. Given a single-view image $\mathbf{I}_t$ and text instruction, $E_{\theta}$ extracts a compact cognition token $\mathbf{z}_t = E_{\theta}(\mathbf{I}_t)$~\cite{li2025vitra}. Conditioned on $\mathbf{z}_t$ and the current kinematic state $s_t$ via AdaLN~\cite{Peebles2022ScalableDM}, $P_{\phi}$ iteratively denoises a sequence of joint-aligned action chunks $(\mathbf{a}_t^i, \dots, \mathbf{a}_{t+N}^i)$. This formulation uses the VLM's scene representation to model complex action distributions in the shared action space.

\noindent{\textbf{Gradient Reversal and Domain Discrimination.}}
To encourage embodiment invariance, the domain discriminator $D_{\psi}$ predicts discrete domain labels $d \in \mathcal{D}$ (e.g., human sensor gloves, dexterous hands, or grippers) from the cognition token $\mathbf{z}_t$ together with the state token $\mathbf{s}_t$. The discriminator interacts with the VLM through a Gradient Reversal Layer (GRL)~\cite{Ganin2014UnsupervisedDA}, which acts as an identity function during the forward pass but multiplies the gradients passed to the VLM by $-\lambda$ during backpropagation. The discriminator learns to classify the embodiment, while the reversed gradient encourages the visual encoder to suppress hardware-specific appearance cues and retain transferable, task-relevant geometric information.

The adversarial branch is used only as a training signal and does not alter the policy interface at inference time. Conditioning the discriminator on the current state helps separate differences that are already explained by kinematics from residual appearance cues in the visual token. Meanwhile, the action objective continues to require information needed for predicting task completion, preventing invariance from being optimized independently of manipulation performance.

\subsubsection{Optimization Objectives}

The framework is trained end-to-end using a joint loss function. The primary task loss is the standard diffusion denoising objective, which minimizes the Mean Squared Error (MSE) of the predicted Gaussian noise $\epsilon$:
$$\mathcal{L}_{\text{MSE}}(\theta, \phi) = \mathbb{E}_{\epsilon \sim \mathcal{N}(0,1), i} \left[ \| \hat{\epsilon}_i - \epsilon \|_2^2 \right],$$
where $\hat{\epsilon}_i$ denotes the noise predicted by the action expert $P_{\phi}$, conditioned on the cognition token $\mathbf{z}_t$ and the hand kinematic state $s_t$.

Concurrently, the domain classification loss is formulated as a standard cross-entropy objective over the predicted domain distribution:
$$\mathcal{L}_{\text{D}}(\psi \mid \theta) = \mathbb{E}_{(\mathbf{I}_t, s_t, d) \sim \mathcal{B}} \left[ - \sum_{k \in \mathcal{D}} \mathbb{I}(d = k) \log D_{\psi}^{(k)}([\mathbf{z}_t, s_t]) \right],$$
where the domain discriminator $D_{\psi}$ explicitly takes the concatenated vector of the cognition token $\mathbf{z}_t$ and the kinematic state $s_t$ as input.

By integrating the GRL, this minimax adversarial game is optimized through a unified final objective function:
$$\mathcal{L}_{\text{final}} = \mathcal{L}_{\text{MSE}}(\theta, \phi) + \lambda \cdot \mathcal{L}_{\text{D}}(\psi \mid \theta),$$
where $\lambda$ controls the strength of the adversarial gradient. This objective encourages the cognition token $\mathbf{z}_t$ to become more domain-invariant, supporting cross-embodiment visual transfer.

\section{Experiments}

\subsection{Training Recipe}

We use a two-stage training pipeline comprising large-scale pre-training for representation alignment and targeted post-training for real-world adaptation. We first pre-train the model on a heterogeneous mixture of OmniShare, VITRA-1M internet videos~\cite{li2025vitra}, and a subset of Open X-Embodiment~\cite{o2024open}, sampled at a 5:4:1 ratio. We then post-train the model on a small set of real-world robot trajectories to adapt it to the target hardware and physical dynamics. To prevent data leakage, the evaluation tasks, objects, and instructions in Table~\ref{tab:unseen_success_rates} are excluded from the robot trajectories used for post-training, with instruction-level non-overlap also verified against VITRA-1M and OXE.

\subsection{Evaluation of Hand-Action Prediction}

Before deploying to the physical robot, we evaluate the hand-motion prediction capability of the pre-trained VLA model on unseen tasks.

\noindent{\textbf{Benchmark.}} We construct an evaluation benchmark with two test splits: OmniShare-Unseen and HOI4D~\cite{liu2022hoi4d}. OmniShare-Unseen contains 200 trajectories involving 20 novel object categories, such as fruits and cardboard boxes, with 3D point-cloud annotations for the hand and interacting objects. To evaluate generalization to unseen environments, we additionally sample 200 trajectories from HOI4D, including tasks such as grasping kettles and disposing of trash.

\noindent{\textbf{Metrics.}} We decompose the evaluation into two temporal phases. All distance-based metrics are reported in millimeters (mm):
\begin{itemize}[leftmargin=*]
    \item \textbf{Pre-grasp Phase ($d_{\text{h-o}}$):} To evaluate the quality and target alignment of the approach motion, we calculate the minimum Euclidean distance between the predicted finger trajectories of the first action chunk and the target object's point cloud. The average initial distances before prediction are 22.4~mm and 20.2~mm for OmniShare-Unseen and HOI4D, respectively.
    \item \textbf{Execution Phase (MPJPE \& MWTE):} For full task execution, we use Mean Per-Joint Position Error (MPJPE) to measure articulation accuracy and Mean Wrist Translation Error (MWTE) to measure global trajectory fidelity.
\end{itemize}

\noindent{\textbf{Performance Analysis.}} As shown in Table~\ref{tab:pretrain}, we compare our framework with VITRA~\cite{li2025vitra}, a VLA model pre-trained on large-scale internet human videos. When trained only on VITRA-1M and OXE (\textbf{w/o OmniShare}), our framework performs comparably to VITRA. Adding OmniShare (\textbf{w/o Adv}) improves performance on both OmniShare-Unseen and HOI4D, demonstrating the benefit of its ground-truth trajectories. Incorporating the domain-adversarial objective (\textbf{Ours}) further improves all metrics, supporting its effectiveness in reducing embodiment-specific visual bias. The two splits capture complementary forms of generalization: OmniShare-Unseen evaluates novel objects within the same acquisition setup, whereas HOI4D introduces shifts in hand appearance, cameras, and environments. Overall, the results show improved hand-motion prediction under visual-domain shifts.

\begin{table}[t]  
    \centering  
    \footnotesize %
    \vspace{-5pt} %
    \caption{Evaluation and ablation of hand action prediction (mm).}  
    \renewcommand{\arraystretch}{0.85} %
    \setlength{\tabcolsep}{3.5pt} %
    \begin{tabular}{l | c c c | c c c}  
        \toprule
        \multirow{2}{*}{Method} 
            & \multicolumn{3}{c|}{OmniShare-Unseen} 
            & \multicolumn{3}{c}{HOI4D} \\ 
             \cmidrule{2-4} \cmidrule{5-7}  
        & $d_{\text{h-o}}$ $\downarrow$ & MPJPE $\downarrow$ & MWTE $\downarrow$ 
        & $d_{\text{h-o}}$ $\downarrow$ & MPJPE $\downarrow$ & MWTE $\downarrow$ \\
         \midrule  
    VITRA~\cite{li2025vitra}  &  20.1 & 16.2 & 14.8 & 16.3 & 13.7 & 11.5 \\
    \midrule
    \textbf{w/o OmniShare} & 19.8 & 15.7 & 14.2 & 15.6 & 14.1 & 11.3 \\
    \textbf{w/o Adv}        & 6.4 & 4.9 & 4.2 & 13.9 & 13.3 & 10.8 \\
    \textbf{Ours}          & \textbf{3.2} & \textbf{2.8} & \textbf{2.5} & \textbf{10.5} & \textbf{7.2} & \textbf{6.1} \\
        \bottomrule
    \end{tabular}  
    \vspace{-10pt} %
    \label{tab:pretrain}  
\end{table}

\subsection{Real-World Dexterous Manipulation}

\noindent{\textbf{Hardware Setup and Data Collection.}}
Our platform features a Paxini Tora robot equipped with 19-DoF DexH13 dexterous hands. Using a teleoperation system, we collected 1,000 real-robot demonstrations across five manipulation tasks. A spatial tracker and a motion-capture glove record 6D wrist poses and fingertip trajectories, which are mapped into robot arm and hand joint configurations through inverse kinematics (IK) retargeting.

\noindent{\textbf{Task Definitions.}} The evaluation suite comprises five tasks: \textbf{Grasp Single Object}, which requires picking up a specified item and placing it into a target box; \textbf{Grasp Multiple Objects}, which involves sequentially moving multiple objects into a receptacle; \textbf{Pour Water}, which transfers liquid from a cup into a bowl; \textbf{Push Cube}, which requires pushing a block to a designated target; and \textbf{Stack Bottle}, which requires placing a bottle securely onto a rack.

\noindent{\textbf{Protocol.}}
We post-train AdvDex, $\pi_{0.5}$, and VITRA using the same robot demonstrations, action representation, and number of training steps. Each method uses a single trained policy and is evaluated over 20 trials per task, with five randomized initial configurations in each of four workspace regions.

\noindent{\textbf{Performance Analysis.}}
Table~\ref{tab:success_rates} compares our framework with the baselines on five seen manipulation tasks. Our full model matches or outperforms the baselines across all tasks. Training from scratch (\textbf{w/o Pre-train}) substantially degrades performance, demonstrating the importance of large-scale pre-training. Removing OmniShare (\textbf{w/o OmniShare}) or the domain-adversarial module (\textbf{w/o Adv}) also reduces success rates, with larger drops on unseen objects and environments. These results highlight the complementary benefits of human demonstrations and adversarial learning, while the 1,000 robot demonstrations adapt the policy to the DexH13 hand and physical dynamics.

\begin{table}[t]
    \centering
    \caption{Success rates on real-world robot dexterous manipulation tasks (in \%).}
    \label{tab:success_rates}
    \renewcommand{\arraystretch}{0.95} %
    \resizebox{\linewidth}{!}{
    \begin{tabular}{l | c c c c c c c}
        \toprule
        \multirow{2}{*}{Method} & 
        \multicolumn{5}{c}{Seen} & 
        \multicolumn{2}{c}{Unseen} \\
        \cmidrule(lr){2-6} \cmidrule(lr){7-8} 
        & 
        \begin{tabular}{@{}c@{}}Grasp \\ Single Object\end{tabular} & 
        \begin{tabular}{@{}c@{}}Multi-Object \\ Grasp\end{tabular} & 
        \begin{tabular}{@{}c@{}}Pour \\ Water\end{tabular} & 
        \begin{tabular}{@{}c@{}}Push \\ Cube\end{tabular} & 
        \begin{tabular}{@{}c@{}}Stack \\ Bottle\end{tabular} & 
        \begin{tabular}{@{}c@{}}Unseen \\ Objects\end{tabular} & 
        \begin{tabular}{@{}c@{}}Unseen \\ Environment\end{tabular} \\
        \midrule
        $\pi_{0.5}$~\cite{intelligence2025pi_} & 75.0 & 60.0 & \textbf{55.0} & 85.0 & 60.0 & 35.0 & 45.0 \\
        VITRA~\cite{li2025vitra}       & 70.0 & 65.0 & 40.0 & 90.0 & \textbf{70.0} & 40.0 & 35.0 \\
        \midrule
        {\scriptsize \emph{Ablations}} &  & & & & & & \\
        ~~~\textbf{w/o Pre-train} & 35.0 & 20.0 & 10.0 & 50.0 & 25.0 & 5.0 & 0.0 \\
        ~~~\textbf{w/o OmniShare} & 75.0 & 55.0 & 45.0 & 85.0 & 65.0 & 25.0 & 20.0 \\
        ~~~\textbf{w/o Adv}       & 70.0 & 60.0 & 40.0 & 80.0 & 50.0 & 15.0 & 30.0 \\
        ~~~\textbf{Ours}          & \textbf{80.0} & \textbf{70.0} & \textbf{55.0} & \textbf{90.0} & \textbf{70.0} & \textbf{50.0} & \textbf{60.0} \\
        \bottomrule
    \end{tabular}
    }
\end{table}

\begin{wraptable}{r}{0.4\textwidth}
    \vspace{-15pt} %
    \centering
    \caption{Success rates on unseen real-world dexterous manipulation tasks via Co-Training (in \%).}
    \label{tab:unseen_success_rates}
    \renewcommand{\arraystretch}{0.95}
    \resizebox{\linewidth}{!}{
    \begin{tabular}{l c c c c}
        \toprule
        \multirow{2}{*}{Method} & 
        \multicolumn{4}{c}{Human Tasks} \\
        \cmidrule(lr){2-5}
        & 
        \begin{tabular}{@{}c@{}}Box \\ Doll\end{tabular} & 
        \begin{tabular}{@{}c@{}}Press \\ Button\end{tabular} & 
        \begin{tabular}{@{}c@{}}Move \\ Bottle\end{tabular} & 
        \begin{tabular}{@{}c@{}}Tool \\ Use\end{tabular} \\
        \midrule
        $\pi_{0.5}$~\cite{intelligence2025pi_} & 30.0 & 50.0 & 30.0 & 0.0 \\
        VITRA~\cite{li2025vitra}       & 40.0 & 45.0 & 25.0 & 20.0 \\
        \midrule
        {\scriptsize \emph{Ablations}} & & & & \\
        ~~~\textbf{w/o Adv (Pre-train)}  & 50.0 & 35.0 & 35.0 & 15.0 \\
        ~~~\textbf{w/o Adv (Post-train)} & 25.0 & 30.0 & 15.0 & 5.0 \\
        ~~~\textbf{Ours}                & \textbf{60.0} & \textbf{70.0} & \textbf{45.0} & \textbf{30.0} \\
        \bottomrule
    \end{tabular}
    }
    \vspace{-10pt} %
\end{wraptable}

\subsection{Zero-Shot Skill Transfer via Co-Training}

A key advantage of our framework is zero-shot transfer of human-demonstrated skills to the robot embodiment. We co-train the policy on 1,000 robot teleoperation trajectories and 1,000 OmniShare human trajectories with mutually exclusive task sets. The evaluation tasks and associated objects appear only in the human subset, requiring the policy to transfer task knowledge from human demonstrations while leveraging robot-specific control experience from other tasks.

We evaluate the co-trained policy on the physical robot across tasks demonstrated only in the human data, including placing a doll into a box, pressing a button, moving a bottle, and using tools. As shown in Table~\ref{tab:unseen_success_rates}, our framework effectively transfers task-relevant knowledge from human demonstrations to the robot. Removing the domain-adversarial objective during either pre-training (\textbf{w/o Adv(Pre-train)}) or post-training (\textbf{w/o Adv(Post-train)}) reduces execution performance, highlighting the benefit of domain-invariant visual representations for human-to-robot skill transfer.

\begin{wrapfigure}{r}{0.3\textwidth} 
    \vspace{-10pt} %
    \centering
    \includegraphics[width=\linewidth]{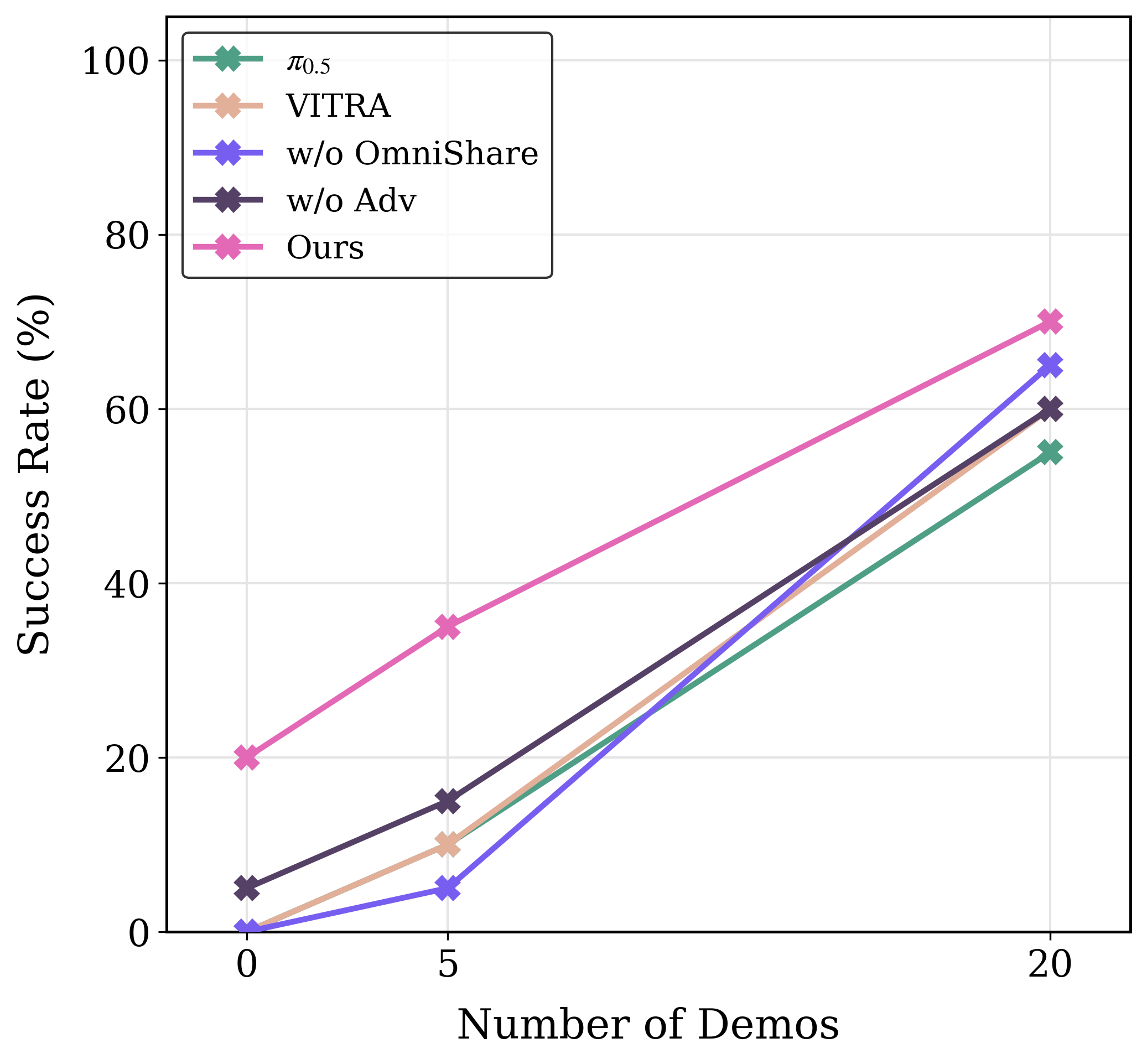}
    \caption{\textbf{Few-shot Fine-tuning Success Rates.} Evaluation on the single-object grasping task across 0, 5, and 20 demonstrations.}
    \label{fig:few_shot}
    \vspace{-10pt} %
\end{wrapfigure}

\subsection{Few-Shot Generalization}

To evaluate data efficiency, we assess the pre-trained model on the single-object grasping task without target-domain fine-tuning and after fine-tuning with five or twenty target-domain demonstrations. As shown in Fig.~\ref{fig:few_shot}, our framework achieves non-zero success in the zero-shot setting, whereas $\pi_{0.5}$ and VITRA fail to complete the task. Performance improves substantially with only five demonstrations. The ablation results further show that removing the domain-adversarial module (\textbf{w/o Adv}) leads to a larger performance drop than excluding OmniShare (\textbf{w/o OmniShare}), highlighting the importance of adversarial learning for data-efficient adaptation.

\begin{wrapfigure}{r}{0.3
\textwidth}
    \vspace{-15pt}
    \centering
    \includegraphics[width=\linewidth]{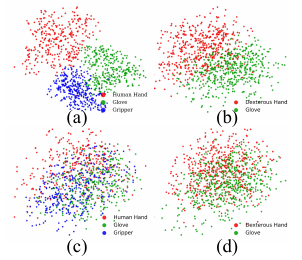}
    \caption{\textbf{t-SNE Feature Distributions.} Intermediate features during pre-training (a) without and (b) with the GRL, and during post-training (c) without and (d) with the GRL.}
    \label{fig:discriminator}
    \vspace{-10pt}
\end{wrapfigure}

\subsection{Effect of Domain-Adversarial Learning}

We visualize intermediate feature distributions using t-SNE (Fig.~\ref{fig:discriminator}). Without the Gradient Reversal Layer (GRL), features from different embodiments form distinct clusters during both pre-training and post-training (Fig.~\ref{fig:discriminator}a, c), indicating embodiment-specific separation in the learned representation. With the GRL, features from different embodiments exhibit substantially greater overlap in the t-SNE embedding (Fig.~\ref{fig:discriminator}b, d). This qualitative result suggests that adversarial training reduces embodiment-specific separation in the learned visual representation.

\section{Conclusions}
We presented AdvDex, a unified Vision-Language-Action framework for learning dexterous manipulation across human and robot embodiments. OmniShare provides large-scale, high-quality human demonstrations; the Joint-Aligned Action Space maps different hand and gripper kinematics into a shared action representation; and domain-adversarial learning reduces embodiment-specific information in the visual features. Real-world experiments demonstrate competitive manipulation performance, improved generalization to unseen objects and environments, and effective human-to-robot skill transfer. These results suggest that combining scalable human demonstrations with a shared action representation and domain-adversarial learning provides a promising path toward more general cross-embodiment manipulation.

\section{Limitations and Future Work}
Although AdvDex aligns actions across embodiments, hardware differences still limit the transfer of fine-grained manipulation skills. Human imitation data alone may not capture the embodiment-specific strategies required for high-precision control. In addition, our current physical evaluation focuses on a single dexterous robot platform, and the shared action representation does not explicitly model embodiment-specific dynamics or contact constraints. Future work could combine the learned policy with reinforcement learning or online adaptation to discover control strategies tailored to each robot and evaluate transfer across additional dexterous platforms.

\clearpage
\medskip
\bibliography{main}

\end{document}